\documentclass[conference]{IEEEtran}
\usepackage{cite}
\usepackage{amsmath,amssymb,amsfonts}
\usepackage{algorithmic}
\usepackage{graphicx}
\usepackage{textcomp}
\usepackage{xcolor}
\def\BibTeX{{\rm B\kern-.05em{\sc i\kern-.025em b}\kern-.08em
    T\kern-.1667em\lower.7ex\hbox{E}\kern-.125emX}}
\usepackage{amssymb}
\usepackage{amsmath}
\usepackage{amsthm}
\usepackage{bbm}
\usepackage{verbatim}

\usepackage{enumitem}
\allowdisplaybreaks

\newlength{\nunitlength}
\newcommand{\al}[1]{\begin{align}#1\end{align}}

\newcommand{\be}{\begin{equation}}
\newcommand{\ee}{\end{equation}}
\newcommand{\beqy}{\begin{eqnarray}}
\newcommand{\eeqy}{\end{eqnarray}}
\newcommand{\beqynn}{\begin{eqnarray*}}
\newcommand{\eeqynn}{\end{eqnarray*}}

\newcommand{\ba}{\begin{array}}
\newcommand{\ea}{\end{array}}
\newcommand{\bmx}{\begin{bmatrix}}
\newcommand{\emx}{\end{bmatrix}}
\newcommand{\bsmx}{\left[\begin{smallmatrix}}
\newcommand{\esmx}{\end{smallmatrix}\right]}
\newcommand{\bmxc}[1]{\left[\begin{array}{@{}#1@{}}}
\newcommand{\emxc}{\end{array}\right]}
\newcommand{\bt}[1]{\begin{tabular}{#1}}
\newcommand{\et}{\end{tabular}}
\newcommand{\bc}{\begin{center}}
\newcommand{\ec}{\end{center}}
\newcommand{\ben}{\begin{enumerate}}
\newcommand{\een}{\end{enumerate}}
\newcommand{\bi}{\begin{itemize}}
\newcommand{\ei}{\end{itemize}}

\newcommand{\Exp}[1]{\mathrm{E}\!\left\{#1 \right\}}
\newcommand{\cov}[1]{\mathrm{cov}\!\left\{#1\right\}}

\renewcommand{\top}{{\mkern-1.5mu\mathsf{T}}}

\newcommand{\diag}{\mathrm{diag}}

\DeclareMathOperator*{\argmin}{argmin}

\newcommand{\Rbb}{{\mathbb{R}}}

\newcommand{\Cbb}{{\mathbb{C}}}

\newcommand{\RN}{\Rbb^{N}}

\newcommand{\calC}{\mathcal{C}}

\newcommand{\calN}{\mathcal{N}}
\newcommand{\calR}{\mathcal{R}}

\newcommand{\calG}{\mathcal{G}}

\newcommand{\calX}{\mathcal{X}}
\newcommand{\calI}{\mathcal{I}}

\newcommand{\A}{\mathbf{A}}

\renewcommand{\H}{\mathbf{H}}
\newcommand{\I}{\mathbf{I}}

\newcommand{\K}{\mathbf{K}}

\newcommand{\T}{\mathbf{T}}
\newcommand{\U}{\mathbf{U}}

\newcommand{\W}{\mathbf{W}}

\newcommand{\Z}{\mathbf{Z}}

\newcommand{\f}{\mathbf{f}}
\newcommand{\g}{\mathbf{g}}
\newcommand{\h}{\mathbf{h}}

\newcommand{\n}{\mathbf{n}}

\newcommand{\s}{\mathbf{s}}

\newcommand{\x}{{\mathbf{x}}}
\newcommand{\y}{{\mathbf{y}}}

\newcommand{\1}{{\mathbf{1}}}

\newcommand{\bmu}{\boldsymbol{\mu}}

\newcommand{\btheta}{\boldsymbol{\theta}}
\newcommand{\bgamma}{\boldsymbol{\gamma}}
\newcommand{\bLambda}{{\boldsymbol{\Lambda}}}
\newcommand{\bSigma}{{\boldsymbol{\Sigma}}}

\newcommand{\0}{{\mathbf{0}}}

\newcommand\ie{\textit{i.e.,}}

\newcommand\iid{\textit{i.i.d.}}

\begin{document}
\newcommand\sitao[1]{\textcolor{red}{[Sitao: #1]}}
\title{\title{GCEPNet: Graph Convolution-Enhanced Expectation Propagation for Massive MIMO Detection} 
% Conference Paper Title*\\

}

\author{\IEEEauthorblockN{Qincheng Lu}
\IEEEauthorblockA{\textit{School of Computer Science} \\
\textit{McGill University}\\
Montreal, Canada \\
qincheng.lu@mail.mcgill.ca}
\and
\IEEEauthorblockN{Sitao Luan}
\IEEEauthorblockA{\textit{McGill University} \\
\textit{Mila - Quebec AI Institute}\\
Montreal, Canada \\
sitao.luan@mail.mcgill.ca}
\and
\IEEEauthorblockN{Xiao-Wen Chang}
\IEEEauthorblockA{\textit{School of Computer Science} \\
\textit{McGill University}\\
Montreal, Canada\\
chang@cs.mcgill.ca}
% \and
% \and
% \and
}

\maketitle

\begin{abstract}
Massive MIMO (multiple-input multiple-output) detection is an important topic in wireless communication and various machine learning based methods have been developed recently for this task. Expectation Propagation (EP) and its variants are widely used for MIMO detection and have achieved the best performance. However, EP-based solvers fail to capture the correlation between unknown variables, leading to a loss of information, and in addition, they are computationally expensive. In this paper, we show that the real-valued system can be modeled as spectral signal convolution on graph, through which the correlation between unknown variables can be captured. Based on such analysis, we propose graph convolution-enhanced expectation propagation (GCEPNet).
GCEPNet incorporates data-dependent attention scores into Chebyshev polynomial for powerful graph convolution with better generalization capacity. It enables a better estimation of the cavity distribution for EP and empirically achieves the state-of-the-art (SOTA) MIMO detection performance with much faster inference speed. To our knowledge, we are the first to shed light on the connection between the system model and graph convolution, and the first to design the data-dependent coefficients for graph convolution.
\end{abstract}

\begin{IEEEkeywords}
massive MIMO detection, machine learning for communication, graph convolution, expectation propagation
\end{IEEEkeywords}

\section{Introduction}

Massive MIMO (multiple-input multiple-output) is a wireless communication technology that uses a very large number of antennas at the base station to significantly increase spectral efficiency and capacity \cite{MarLYN16}. One of the outstanding challenges in massive MIMO is to efficiently detect the transmitted signal. % \cite{Neev2019learning}. 
Finding the optimal detector can be modeled as solving an integer least square problem,  which is NP-hard \cite{Ver89}. As the complexity of the traditional methods for near-optimal solutions still grows prohibitively as the number of users increases, especially for large-scale problem, methods inspired by deep learning have emerged as promising solutions to meet the high throughput and low latency requirements \cite{Nhan2021application}. Achieving a comparable detection performance to the traditional methods,  the learning-based methods shift the significant part of computational cost to an offline training phase and significantly shorten run time in the online detection (inference) phase \cite{Ly2022Leverage}.

For improving the performance of conventional detectors \cite{BaiCY14, ChaCW22, WenC17, ChaCX20, cespedes2014expectation, donoho2009message},
various machine learning methods for MIMO detection
%learning-based MIMO detection methods 
have been proposed \cite{Ly2022Leverage},
including 
pure learning-based methods \cite{Scotti2020graph, pratik2020re},
and the methods which incorporate learnable parameters  \cite{he2018model, Neev2019learning, Nhan2020deep} 
or learnable modules \cite{kosasih2022graph, kosasih2022graph2, he2023gnn, liu2023model} into some conventional detectors.
%\red{Why don't you consider [14]-[17]?}
% \red{1. If learning is used to aid a main method, I won't say it's 
% a learning-based method. 
% 2. Do all these papers use learning methods to aid message passing detectors
% or some do and some don't? If it's the latter, state clearly about this
% and give an explanation about what MP is and state why they need 
% learning }
Among them, graph neural network (GNN) aided massage passing (MP) detectors \cite{kosasih2022graph, kosasih2022graph2, he2023gnn, liu2023model} 
stands out with better performance. %detection performance. 
% \blue{17 does not change GEPNet's architecture, they propose a trick in updating parameters in EP, which also works well in GEPNet.}
% \red{Is IEP-GNN [17] better than GEPNet?}
% \blue{17 is better than GEPNet, but the efficiency is the same because they still use the same GNN as GEPNet. IEP-GNN is not open-sourced, so I cannot do comparison.}
% \red{When you talk}
% \blue{GPICNet use the GNN in GEPNet, it replace EP in GEPNet by BPIC, it is more efficient but the performance is bad. }
% \red{State that in the paper}
% \blue{I will re-write this paragraph}
In GNN aided MP detectors, 
GNN is used to improve the approximation of posterior distribution in 
various MP detectors, including Expectation Propagation (EP) \cite{cespedes2014expectation}, Approximate Message Passing (AMP) \cite{donoho2009message} and Bayesian parallel interference cancellation (BPIC) \cite{kosasih2021bayesian}.
The GNN aided EP framework (GEPNet) \cite{kosasih2022graph} and its variant IEP-GNN \cite{liu2023model} (to be elaborated later) achieve  the best performance among well-known conventional sub-optimal detectors and 
those which apply machine learning techniques.

Here is a brief history about the incorporation of GNNs into MP.
GEPNet is the pioneering work, but its computation involves matrix inversion 
%in each neural network layers 
due to the EP iterations \cite{kosasih2022graph}.
As it is inefficient, later efforts were made to %enhance its efficiency by 
leverage other MP methods in GEPNet. 
For example, GPICNet \cite{kosasih2022graph2} substitutes the EP module with BPIC,
and AMP-GNN \cite{he2023gnn} replaces it by AMP.
Although both changes reduce complexity, 
they degrade GEPNet's performance.
Very recently, IEP-GNN \cite{liu2023model} was proposed to
improve the computation of local posterior information in the EP iteration,
which enhances the performance of GEPNet while maintaining the same computational cost.
However, to our best knowledge, there is no existing work that aims to improve the efficiency of the GNN module in GEPNet. In this paper, we focus on addressing the inefficiency issue of the GNN module in GEPNet, with no performance degradation and sometimes even performance enhancement. The proposed method can be seamlessly integrated with IEP-GNN.

% \red{Confusing. What's the relation between GEPNet [13] and GPICNet [13]?
% Is GEPNet the best one among all available methods which use learning?}
% \blue{Yes}
Meanwhile, the theoretical justification, expressivity, and efficiency of the currently widely adopted GNNs framework for MIMO detection are under-explored. For example, existing GNNs for MIMO are intuitively based on the pair-wise Markov Random Field (MRF) factorization of the posterior \cite{Scotti2020graph, kosasih2022graph, kosasih2022graph2}, where a spatial GNN is utilized to parameterize and learn the representation of the variable and factor nodes in the MRF model. However, the formulation of these GNNs inherits from works that combine deep learning with MRF \cite{Scotti2020graph}, but less motivated by the unique property of MIMO detection problem. In addition, spatial GNNs originate as the low order approximation of the graph convolution \cite{Defferrard2016}, which loses high-order topology information. To enhance the expressivity of the MIMO-specific GNNs, we investigate the %theoretical 
connection between the system model and the graph convolution, and propose a novel
%\red{Is it appropriate to add the word "novel" here?}
graph convolution-enhanced EP detector (GCEPNet) \footnote{Code available at: https://github.com/wzzlcss/GCEPNet}, which achieves state-of-the-art detection performance with fewer training parameters and lower inference complexity.

The rest of the paper is organized as follows. Section~\ref{sec:preliminaries} gives notation and reviews spectral graph convolution. Section~\ref{sec:sysmodel} introduces the MIMO system model. 
The machine learning approach for MIMO detection is outlined in Section~\ref{sec:mlmimo}.
Section~\ref{sec:ep} presents the use of GNN in the EP detector. We propose the GCEPNet in Section~\ref{sec:gcepnet} and show its efficiency in Section~\ref{sec:gepnet-compare}.
Section~\ref{sec:exp} provides training details and performance comparison. Section~\ref{sec:conclusion} concludes the paper.

% \red{As there is space, state how the paper is organized.}
% \red{If it can help to explain Fig 2, add it}

\section{Preliminaries}

\label{sec:preliminaries}

\subsection{Notation}
%First we introduce notation to be used through the paper.
We use bold capital and small letters for matrices and vectors,
respectively. The $(i,j)$ entry of $\A$ is denoted by $\A(i,j)$ or $a_{ij}$,
the $i^{th}$ row of $\A$ is denoted by $\A_{i,:}$,
and the $i^{th}$ entry of $\s$ is denoted by $\s(i)$ or $s_i$.
We denote $\1=[1,\ldots,1]^\top$.
When a real-valued random vector $\x$ follows the normal distribution
with mean $\bmu$ and covariance matrix $\bSigma$, we write
$\x\sim \calN(\bmu,\bSigma)$ and use $N(\x\!:\!\bmu,\bSigma)$ to denote
its  density function.
% We use $[B||C]$ and $[B, C]$ to denote column and row concatenation of matrices B and C, respectively.

\subsection{Graph and Convolution}
For a graph ${\cal G} = ({\cal V},  {\cal E})$, where ${\cal V}$  is the  vertex set  with $N$ vertices (or nodes),  \ie{} $|{\cal V}| = N$, and ${\cal E}$ is the set of edges, the adjacency matrix of ${\cal G}$ is defined as $\mathbf{A} \in \Rbb^{N\times N}$ with $a_{ij} = 1$ if $e_{ij} \in {\cal E}$ and $a_{ij} = 0$ otherwise. The degree matrix is denoted as $\mathbf{D} = \text{diag}(d_{ii}) \in \Rbb^{N\times N}$ with $d_{ii} = \sum_{j} a_{ij}$. There are various graph  Laplacians in the literature \cite{luan2022revisiting, lu2024representation}, %\red{Give a reference here}
and among them, the symmetric normalized graph Laplacian is most widely used one $\mathbf{L} = \mathbf{I} - \mathbf{D}^{-\frac{1}{2}} \mathbf{A} \mathbf{D}^{-\frac{1}{2}}$. Let its eigen-decomposition be $\mathbf{L} = \mathbf{U} \bLambda \mathbf{U}^\top$, where $\bLambda = \diag(\lambda_i)$ with $0 = \lambda_1 < \lambda_2 \leq \cdots \leq \lambda_N \leq 2$ and $\U$ is orthogonal. Given a graph signal $\s \in \RN$, the graph Fourier transform and the inverse graph Fourier transform are defined as 
\begin{equation}
\hat{\s} := {\cal F}(\s) := \U^\top \s, \quad \s:={\cal F}^{-1}(\hat{\s}) := \U \hat{\s},
\label{eq:gf}
\end{equation}
respectively.
The graph convolution between graph signals 
$\s_1 \in \Rbb^N$ and $\s_2 \in \Rbb^N$ 
%, denoted by $*_{{\cal G}}: \mathbb{R}^N \rightarrow \RN$, 
is defined as (see, e.g., \cite{Defferrard2016})
\begin{equation}
\hspace*{-2mm}
\s_2 *_{{\cal G}} \s_1 := {\cal F}^{-1} \left( {\cal F}(\s_2) \odot {\cal F}(\s_1)\right) \!=\! \U \left( \U^\top \s_2 \odot \U^\top \s_1 \right), 
\label{eq:gcv}    
\end{equation}
where $\odot$ is an element-wise multiplication operation.

\subsection{Spectral graph convolution}
We briefly review the spectral graph convolution for graph signal processing and ChebNet  \cite{Defferrard2016}, 
%\red{Do we need all of these references?}
which will be used later.
Suppose a function $g_{\btheta}$ which operates on the eigenvalues of the graph Laplacian $\mathbf{L}$
is defined as $\g_{\btheta}(\lambda(\mathbf{L})) := [g_{\btheta}(\lambda_1), \cdots, g_{\btheta}(\lambda_N)]^\top$.
The graph spectral filter associated with $\g_{\btheta}(\lambda(\mathbf{L}))$ is a linear operator $\T_{g_{\btheta}}: \RN \rightarrow \RN$ satisfying
\begin{equation}
\T_{g_{\btheta}} \s = {\cal F}^{-1} \left( \g_{\btheta}(\lambda(\mathbf{L})) \odot {\cal F}(\s)\right) = \U \mbox{diag} \left( \g_{\btheta}(\lambda(\mathbf{L})) \right) \U^\top \s.
\label{eq:gcf1}
\end{equation}
Various designs of $g_{\btheta}(\lambda)$ lead to different spectral graph convolutions. It is proved that any graph convolution with well-defined analytic spectral filter can be written in a truncated block Krylov form \cite{luan2019break} and polynomial filter is one of the most commonly used instances. For a polynomial $g_{\btheta}(\lambda) = \sum_{m=0}^{M} c_m \lambda^m$,
%with coefficients $w_m$,
the graph convolution in \eqref{eq:gcf1} becomes
\begin{equation}
\T_{g_{\btheta}} \s = \U \left( \sum_{m=0}^{M} c_m \bLambda^m \right) \U^\top \s = \sum_{m=0}^{M} c_m \mathbf{L}^m \s.
\label{eq:gc-poly}
\end{equation}
% \red{Why is this an approximation?}
% \blue{Because here we use the polynomial to approximate some unknown g}
%When $M$ is large, computing \eqref{eq:gc-poly} suffers from overflow. 
In \cite{Defferrard2016}, ChebNet is proposed to use Chebyshev polynomials to fit \eqref{eq:gc-poly} as
\begin{equation}
\label{eq:chebyshev_polynomial}
\T_{g_{\btheta}} \s = \sum_{m=0}^{M} w_m T_m(\hat{\mathbf{L}}) \s,
\end{equation}
where $T_m(x)$ are Chebyshev polynomials defined by
% where $\hat{\mathbf{L}} = 2 \mathbf{L} / \lambda_{\text{max}}( \mathbf{L}) - \I$, 
% and the Chebyshev polynomials are defined by 
$T_m(x) = 2 x T_{m-1}(x) - T_{m-2}(x)$, $T_0(x) = 1$ and $T_1(x) = x$.

\section{System Model}

\label{sec:sysmodel}

Suppose that in a MIMO system, the transmitter is equipped with $N_t$ antennas and the receiver is equipped with $N_r$ antennas, and the transmitted signal $\x_c$ and the received signal $\y_c$ satisfy the following model:
\begin{equation}
\y_c = \H_c \x_c + \n_c,\label{eq:sys-model}
\end{equation}
where the channel matrix $\H_c \in \Cbb^{N_r \times N_t}$ is assumed to be known to the receiver; 
the elements of $\x_c \in \calX^{N_t}_c$ are independently distributed over the constellation set $\widetilde{\calX}_k= \{ k_1 + k_2 j: k_1, k_2 = \pm 1, \pm 3, \cdots, \pm (2^k -3), \pm (2^k -1)\}$, $k=1, 2, 3$ correspond to QPSK, \ie{} 4QAM, 16QAM, 64QAM constellations, respectively; and the complex circular Gaussian noise $\n_c \sim \calC \calN (\mathbf{0}, \sigma_c^2 \I)$.
% with complex normal distribution $\calC \calN (\mathbf{0}, \sigma_c^2 \I)$
%has its real and imaginary parts independently follows $\calN(\mathbf{0}, \frac{1}{2} \sigma_c^2 \I)$).
% We assume perfect channel state information (CSI), i.e.,
% the channel matrix $\H_c \in \Cbb^{N_r \times N_t}$, is known at the receiver.
Given $\y_c$ and $\H_c$. our objective is to estimate $\x_c$.
% The optimal solution found by the maximum likelihood (ML) detector
% solves the integer least squares (ILS) problem: 
% \begin{equation}
% \min_{\x_c \in \widetilde{\calX}_k^{N_t}} \| \y_c - \H_c \x_c \|_2^2.\label{eq:ils}
% \end{equation}
Define %the following matrices to simplify the complex operations,
\al{
&\y = \begin{bmatrix} \calR(\y_c) \\ \calI(\y_c) \end{bmatrix}, \
\H = \begin{bmatrix} \calR(\H_c) & -\calI(\H_c) \\ \calI(\H_c) & \calR(\H_c) \end{bmatrix}, \ \x = \begin{bmatrix} \calR(\x_c) \\ \calI(\x_c) \end{bmatrix}, \nonumber \\
&\n = \begin{bmatrix} \calR(\n_c) \\ \calI(\n_c) \end{bmatrix}, \ 
\calX_k = \{ \pm 1, \cdots, \pm (2^k -1) \}, \label{def:ils-real}
}
where $\calR(\cdot)$ and $\calI(\cdot)$ denote the real and imaginary parts. Then the complex system model in \eqref{eq:sys-model} becomes
\begin{equation} 
\y = \H \x + \n, \ \ 
\n \sim \calN(\0, \sigma_n^2\I). \label{eq:sys-model-real}
\end{equation}
where $\x\in \calX_k^{2N_t}, \H\in \Rbb^{2N_r\times 2N_t}$, $\sigma_n^2=\sigma_c^2/2$.
%which is also referred to as the forward model \cite{pratik2020re}.
% By Baye's rule,
% the posterior distribution of $\x^*$ is
% \begin{equation}
% p(\x | \y, \H) \propto p(\y | \x, \H) p(\x) \propto \text{exp}\{ - \| \y - \H \x \|_2^2 \} p(\x).\label{eq:post}
% \end{equation}
% The goal of MIMO detection is to solve the MAP problem:
% \begin{equation}
% \underset{\x \in \calX^{2 N_t}_k}{\text{max}} p(\x | \y, \H).\label{eq:map}
% \end{equation}
% \red{Will \eqref{eq:post} and \eqref{eq:map}  be used later?
% If not, comment out them.}
The maximum likelihood method for detecting $\x$ in \eqref{eq:sys-model-real}
solves the integer least squares (ILS) problem:
\begin{equation} 
\underset{\x \in \calX^{2 N_t}_k}{\text{min}} \| \y - \H \x \|_2^2.\label{eq:ils-real}
\end{equation}
% When $p(\x)$ is non-informative,
% \eqref{eq:map} reduces to \eqref{eq:ils-real}.
% \red{Don't follow the writing}
% When there are more transimitting antennas than receiving antennas ($\N_t>\N_r$) in the communication system,
% the problem in \eqref{eq:ils-real} is referred to as an underdetermined MIMO problem.
%When $\H$ is a square matrix with full column rank,
%it is called an determined MIMO problem.
In this paper we assume that $\H$  has full column rank.
This is usually true in practice.
% Algorithms that searches for the optimal solution have been proposed for the overdetermined MIMO problem \cite{},
% and have been extended to the underdetermined cases \cite{}. 
%While solving \eqref{eq:ils-real} is NP-hard, 
% and the complexity of all existing algorithms for solving it grows exponentially with the number of $\N_t$ for the worst case.
As \eqref{eq:ils-real} is NP-hard, the maximum likelihood detection 
for a large MIMO system is prohibited.

%\addtolength{\topmargin}{+0.1cm}

\section{Machine learning for MIMO Detection}

\label{sec:mlmimo}

From the machine learning perspective, MIMO detection can be regard as a classification problem, where we classify each component of the unknown $\x$ in \eqref{eq:sys-model-real} into $2^k$ categories defined by the constellation set $\calX_k$. For convenience, we abuse the notation and regard $\calX_k$ as a tuple, and order its $2^k$ elements, with the $j^{th}$ 
%\red{$c$-th is also used in the paper. Take one for consistency.}
element denoted by $\calX_k(j)$.
% Let $\bxi \in \mathbb{Z}^{2^k}$ denote the vector form of $\calX_k$,
% i.e., the elements of $\bxi$ are the elements of $\calX_k$. 
Then, our goal is to learn a function $\f_{\btheta}: \y, \H \xrightarrow{} \Z \in \Rbb^{2N_t \times 2^k}$, where $\Z(i, j)$ is proportional to the probability of $x_i$ being labeled as $\calX_k(j)$. 
And $\btheta$ is a vector of trainable parameters of the network, which is to be optimized by minimizing a loss function with training data samples. %$\{ (\x^{(i)}, \y^{(i)}, \H^{(i)}) \}_i$. 
In application, $\f_{\btheta}$ computes the soft estimation $\hat{a}_{i}$ of $x_i$ and its variance $\hat{b}_i^2$:
\begin{align}
&\text{Pr}\big(x_i = \calX_k(j)|\y,\H\big)%p(x_n=\calX_k(c)|\y,\H) 
%= \text{softmax} (\Z(n, :))
= \frac{\text{exp}\{\Z(i, j)\}}{\sum_{\ell} \text{exp} \{\Z(i, \ell)\}},\label{eq: softmax}\\
&\hat{a}_{i} = \textstyle  \sum_{j} \calX_k(j) \times \text{Pr}\big(x_i = \calX_k(j)|\y,\H\big),  
\label{eq: softestimate}\\
&\hat{b}_i^2 = \textstyle \sum_{j} \left( \calX_k(j) - \hat{x}_{i} \right)^2 \times \text{Pr}\big(x_i = \calX_k(j)|\y,\H\big)
\label{eq: varianceestimate}
\end{align}
%\sitao{$\calX_k(c)$ is a set instead of a vector, probably use $x_n(c)$ or redefine a vector}
where \eqref{eq: softmax} is referred to as the softmax normalization of 
the $i^{th}$ row of $\Z$. 
A component-wise hard decision is obtained by rounding each element of $\hat{a}_{i}$ independently onto %$\calX_k$:
the constellation set:
\begin{equation}
\hat{x}_{i} = \argmin_{x \in \calX_k} | x - \hat{a}_{i} |.
\label{eq: round}
\end{equation}
% \red{1. Is $\Z_n$ the $n$-th column of $\Z$? If yes, change it to $\Z(:,n)$
% 2. Is this a framework used in all the relevant literature?
% Cite references.}

\section{GNN-aided Expectation Propagation}

\label{sec:ep}

% \red{This section can be shortened if there is a space problem.}
%We introduce the motivation of incorporating GNNs into the EP detector.
We briefly introduce the EP detector
and explain the use of GNN in EP.
For the system model defined in \eqref{eq:sys-model-real},
the posterior probability density of $\x$ is given by Bayes' rule:
\begin{equation}
\hspace*{-2mm}
p(\x | \y, \H) \! \propto \! p(\y|\x, \H) p(\x) \!=\! N(\y\!:\! \H \x, \sigma_n^2 \I) \prod_{i=1}^{2 N_t} p(x_i),
\label{eq:sys-model-bayes}
\end{equation}
where the prior $p(x_i) = \mathbbm{1}_{x_i \in \calX_k}$,
%$ \frac{1}{| \calX_k|} \sum_{x \in \calX_k} \delta(x_i - x)$ 
which is an indicator function.
The EP detector tries to find an approximate Gaussian distribution $q(\x | \y, \H) =N(\x\!:\!\bmu_{\x}, \bSigma_{\x})$ to estimate the intractable $p(\x | \y, \H)$,
% \red{If $\x_i^*$ is uniformly distributed, why is it intractable?
% Does the literature assume the distribution of $\x_i^*$ is unknown
% or not uniform or Gaussian when they employ EP? 
% If it's true, a change is needed in Sec III.}
% \blue{Assuming uniform distribution gives the MAP problem, which is of exponential complexity to solve. Therefore, there they use a Gaussian to replace the uniform prior so that the posterior is also a Gaussian.}
% \red{1. If "intractable" is in that sense, the problem is still there
% even if $\x^*$ is Gaussian, as you would have a regularized ILS problem.
% 2. If $\x^*$ is uniform, $p(x_i)$ is a constant.
% Then the RHS of (9) is just Gaussian.
% }
% \blue{If the prior is uniform, then the posterior is $$p(\x|\y) = \frac{1}{Z} \exp(-\|\y-\H\x \|^2/2\sigma^2) $$ where $Z$ normalizes it such that $\sum_{\x \in \calX_k^n} p(\x|\y) = 1$. We call this the true posterior distribution, it is intractable because of $Z$. The approach here want to use $\cal{N}(\bmu_{\x}, \bSigma_{\x})$ to approximate it where we can calculate $\bmu_{\x}$ exactly.}
% \red{$p(\x|\y)=p(\y|\x)p(\x)/p(\y)$. 
% For MAP, $p(\y)$ is not relevant to $\x$,
% thus $\max p(\x|\y)$ is equivalent to $\max p(\y|\x)p(\x)$.
% If we just want to get $p(\x|\y)$ exactly or approximately,
% then  (9) is irrelevant.
% Also the distribution of $\x^*$ is irrelevant?
% }
% \blue{if we consider (9) then we need to solve (6), here we don't want to solve (6) so we modify (9) by modifying the prior.}
and then rounds $\bmu_{\x}$ to the nearest integer point in the constellation set $\calX_k^n$ to get the estimation of $\x$ in \eqref{eq:sys-model-real}. To this end, we need first to find out a suitable prior $p(x_i)$, and the EP detector replaces the non-Gaussian prior in \eqref{eq:sys-model-bayes} with unnormalized Gaussian distribution as follows,
%containing unknown parameters $\boldsymbol{\Theta} \in \Rbb_{+}^{2N_t \times 2N_t}$ and $\boldsymbol{\gamma} \in \Rbb^{2N_t}$
\begin{align}
\prod_{i=1}^{2 N_t} p(x_i) 
& \approx 
\prod_{i=1}^{2 N_t} g(x_i) 
:= \prod_{i=1}^{2 N_t} \exp\big\{\!-\frac{1}{2} \theta_{ii} x_i^2 +  {\gamma}_i x_i  \big\} \\
& =\exp\big\{\!-\frac{1}{2} \x^\top \boldsymbol{\Theta} \x + \x^\top \boldsymbol{\gamma}  \big\},
\end{align}
where $\boldsymbol{\Theta}=\diag(\theta_{ii})$ and
$\bgamma=(\gamma_i)$.
% We also denote the approximation to $p(x_i)$ as $g(x_i) = \exp\{-\frac{1}{2} \boldsymbol{\Theta}_{ii} x_i^2 +  \boldsymbol{\gamma}_i x_i  \}$.
The EP detector finds $\boldsymbol{\Theta}$ and $\bgamma$ through iterative methods \cite{cespedes2014expectation}.
At the $t^{th}$ iteration,
the posterior approximation $p^{(t)}(\x | \y, \H)$ is given by
\begin{align*}
&p^{(t)}(\x | \y, \H) = N(\x\!:\!\bmu_{\x}^{(t)}, \bSigma_{\x}^{(t)}) \nonumber \\
&\propto N(\y\!:\!\H \x, \sigma_n^2 \I) \exp\{-\frac{1}{2} \x^\top \boldsymbol{\Theta}^{(t-1)}\x + \x^\top \boldsymbol{\gamma}^{(t-1)} \}.
\end{align*}
% It is easy to verify that \blue{Do we need this detail?}
% \begin{align*}
% \bSigma_{\x}^{(t)} &= (\sigma_n^{-2} \H^\top \H + \boldsymbol{\Theta}^{(t-1)})^{-1}, \nonumber \\
% \bmu_{\x}^{(t)} &= \bSigma_{\x}^{(t)}(\sigma_n^{-2} \H^\top \y + \boldsymbol{\gamma}^{(t-1)}).
% %\label{eq:ep_1}
% \end{align*}
% Note that the initialization $\boldsymbol{\Theta}^{(0)} = \sigma^{-2}_{\x} \I$, $\boldsymbol{\gamma}^{(0)} = \mathbf{0}$ gives the minimum mean square error (MMSE)
% estimate $\bmu_{\x}^{(1)}$ of $\x$ in \eqref{eq:sys-model-real}.
We can take $\boldsymbol{\Theta}^{(0)} = \cov{\x}= \sigma^{-2}_{\x} \I$
(suppose $\sigma_{\x}$ is known) and $\boldsymbol{\gamma}^{(0)} = \mathbf{0}$.
Denote the $i^{th}$ marginal of $p^{(t)}(\x | \y, \H)$ as 
\begin{equation}
q^{(t)}(x_i | \y, \H) = N (\bmu_{\x}^{(t)}(i), \bSigma_{\x}^{(t)}(i, i)).
\label{eq:iga-1}
\end{equation}
% To reduce the computational complexity, 
% the EP detector further approximates $p^{(t)}(\x | \y, \H)$  
% by the product of %individual distributions:
% local Gaussian distributions $q^{(t)}(x_i | \y, \H)$:
% \begin{equation}
% p^{(t)}(\x | \y, \H)  
% \approx 
% q^{(t)}(\x | \y, \H) 
% :=
% \prod_{i=1}^{2 N_t} q^{(t)}(x_i | \y, \H). \label{eq:iga-2}
% \end{equation}
%\red{State what $ q^{(t)}(x_i | \y, \H)$ is.}
%Now we show how to compute $\bTheta^{(t)}$ and $\bgamma^{(t)}$.
At each iteration, 
an improved posterior $\hat{p}^{(t)}(x_i)$ 
of $x_i$ can be obtained 
by replacing the approximation factor $g(x_i)$ from $q^{(t)}(x_i | \y, \H)$ by the true factor $p(x_i)=\mathbbm{1}_{x_i \in \calX_k}$ \cite{cespedes2014expectation}:
%the $i$-th cavity density of $q^{(t)}(\x | \y, \H)$, denoted as $q^{\backslash i}(x_i)$,
%is obtained by removing the approximation factor $g(x_i)$ from $q^{(t)}(x_i | \y, \H)$.
%And an improved posterior of $x_i$, denoted as $\hat{p}^{(t)}(x_i)$,
%is calculated by replacing $g(x_i)$ by the true factor $p(x_i)$ in $q^{(t)}(x_i | \y, \H)$ as follows
\begin{equation}
q^{(t) \backslash i}(x_i) := \frac{q^{(t)}(x_i | \y, \H)}{\exp\{-\frac{1}{2}\theta_{ii}^{(t-1)} x_i^2 + {\gamma}_i^{(t-1)} x_i \}}, 
\label{eq:qti}
\end{equation}
\begin{equation}
\hat{p}^{(t)}(x_i) \propto q^{(t) \backslash i}(x_i) \mathbbm{1}_{x_i \in \calX_k}.
\label{eq:epestimation}
\end{equation}
From \eqref{eq:qti} and \eqref{eq:epestimation}, the EP detector finds the cavity distribution
$q^{(t) \backslash i}(x_i) = N(x_i\!:\! a^{(t)}_i, (b_i^{(t)})^2)$
and $\hat{p}^{(t)}(x_i) = N(x_i\!:\! \hat{\mu}^{(t)}_i, (\hat{\sigma}^{(t)}_i)^2)$ as shown in \cite{cespedes2014expectation}.
Then the EP detector finds  $\theta^{(t)}_{ii}$ and $\gamma^{(t)}_i$ 
satisfying
\begin{equation}
N (\hat{\mu}^{(t)}_i, \hat{\sigma}^{(t) 2}_i) \propto q^{ (t) \backslash i}(x_i) \exp\{-\frac{1}{2} \theta^{(t)}_{ii} x_i^2 + \gamma_i^{(t)} x_i \}.
\label{eq:ep-update}
\end{equation}
Although the EP detector has been shown to outperform various conventional sub-optimal detectors across different problem sizes and QAM orders \cite{cespedes2014expectation}, it only uses the local Gaussian distribution \eqref{eq:iga-1} to define the cavity distribution in \eqref{eq:qti}, which treats each element in $\x$ independently. Such assumption of independence is strong, ignoring the correlation information in the off-diagonal elements of $\bSigma_{\x}^{(t)}$, and thus failing to capture the synergy between the components of $\x$.
% \sitao{EP treats each element in $\x$ independently (or put this strong assumption in $\x$), failing to capture the dependency between them and thus, lose important information for modeling, is my statement correct?}\blue{Yes}
% the approximation in \eqref{eq:iga-2} might cause information loss concerning the potential correlations among each $x_i$.
% \red{Where is (18) used?  (12) needs (19).}
% \blue{(18) says the indenpendent assumption, we will refer to this assumption later}
% \red{EP uses this important assumption. This section introduces EP and it has to show where this assumption is used. Otherwise there will be a problem to introduce
% GEPNet.}
% \blue{OK, i will add a section after this section to talk about how to address the limitation in (18)}
% \red{Don't need to introduce a new section. Here the problem is not about how to address the limitation, it's about where the assumption is used.}
% \blue{Previously we give the expression of $\bSigma_{\x}$, it has covariance there since we will have non-zeros in off-diagonals. In (18) and (19) we won't have nonzero off-diagonals, OK Let's have a quick chat.}
To deal with this issue, 
GEPNet uses a graph neural network to model the such correlations and learn
$N(\hat{a}_i^{(t)}, (\hat{b}^{(t)}_i)^2)$ 
% \red{Check closely} \blue{what symbol should I use?} 
% \blue{20 is not related to GNN, GNN only do some thing for 19, after that, it has nothing to do with GNN}
% \red{Need to}
% \red{But our notation has to be consistent.
% Compare it with (20). Make sure a reader won't be confused.}
via \eqref{eq: softmax}, \eqref{eq: softestimate}, \eqref{eq: varianceestimate}
to replace the $q^{(t) \backslash i}(x_i)$ in \eqref{eq:epestimation} at the $t^{th}$ EP iteration.
and use $\hat{a}_i^{(T)}$ at the last EP iteration ($t = T$) to give the final detection in \eqref{eq: round}. 
% \sitao{So it's a GNN combined with $T$ EP iterations?}\blue{GNN is combined with EP in every iteration, the final output is from GNN}
% \red{Is it $q^{(t)\backslash i}(x_i)$ for all $t$?}
% \blue{Yes, do this for all t}
% Thus %takes such correlations into account and 
% effectively improve the posterior distribution approximation. % consequently.
More details about GEPNet will be given in Section~\ref{sec:gepnet-compare}.

\addtolength{\topmargin}{0.02 in}

\section{Efficient Graph Convolution-Based MIMO Detector}

\label{sec:gcepnet}

In this section,
we propose the graph convolution-enhanced expectation propagation (GCEPNet), a novel graph neural network architecture with theoretical support that provides an efficient scheme to incorporate correlations among unknown symbols for MIMO detection. The proposed method effectively addresses not only the limitation of 
the loss of correlation information
% the independent Gaussian assumption \red{We need to discuss this writing} 
in the original EP detector but also the computation bottleneck in the existing state-of-the-art GNNs for MIMO, and lays the foundation for designing better graph convolution neural networks for   MIMO detection in future research.
%\red{I don't see any theory. Rewrite this}

\begin{figure*}
\centering
\includegraphics[width=0.99\textwidth]{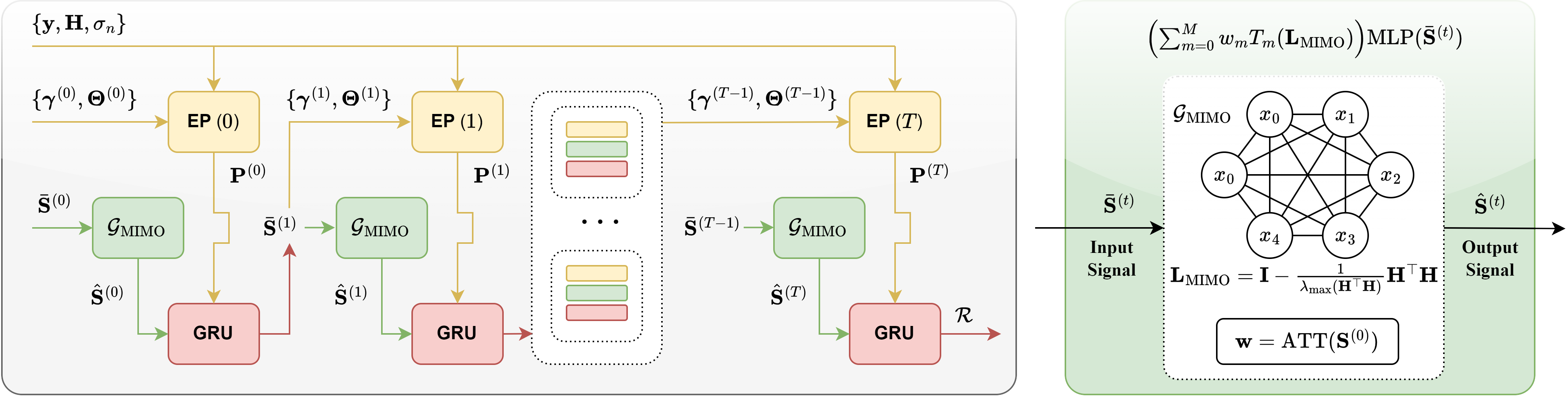}
\caption{The structure of GCEPNet.
The left panel shows the main workflow, where an iteration $t$ contains three modules that perform EP calculation, graph convolution and GRU gating respectively. Arrows indicate the data flow.
The right panel illustrates the graph convolution process,
that computes \eqref{eq: hidden-feature} and \eqref{eq: gcepnet},
using $N_t = 3$ as an example.}  
\label{fig:GCEPNet}
\end{figure*}

\subsection{The Graph Convolution Form of the System Model}
In this section,
we  show that the system model can be described by a graph convolution.
In order to learn the representation of unknown $x_i$ as well as the correlations among them,
for each instance,
we define a fully connected self-loop graph 
${\cal G}_{\text{MIMO}} = (\mathcal{V}_{\x}, \mathcal{E}_{\x})$ 
%with $|\mathcal{V}_{\x}| = 2N_t$.
%In ${\cal G}_{\text{MIMO}}$, 
with node $i\in \mathcal{V}_{\x}$ representing $x_i$  from the problem instance.
In ${\cal G}_{\text{MIMO}}$,
there exists edge $e_{ij}$ between each pair of nodes $(i,j)$, 
indicating the potential correlations.
The available $\y$, $\H$ and $\sigma_n$ of the instance are encoded in the Laplacian and graph signal on ${\cal G}_{\text{MIMO}}$.
We define a   normalized graph Laplacian of $\calG_{\text{MIMO}}$ as
\begin{equation}
\mathbf{L}_{\text{MIMO}} = \I - \alpha \H^\top \H := \I - \W, 
\ \ \alpha:=1/\lambda_{\text{max}}(\H^\top \H),
\label{eq:L-MIMO}
\end{equation}
where $\W=\alpha\H^\top \H$ is a weight matrix of the graph
and $\W(i,j)$ is a weight for the edge $e_{ij}$.
Like different variants of the graph Laplacian \cite{lu2024representation}, 
$\mathbf{L}_{\text{MIMO}}$ has nonnegative eigenvalues,
more specifically, $\lambda(\mathbf{L}_{\text{MIMO}}) \in [0, 1)$. 
Note that the weights $\W(i,j)$ can be negative for $i \neq j$, and for self-loop edges, $\W(i,i)$ are not all equal. Thus, $\mathbf{L}_{\text{MIMO}}$ has some differences with the existing graph Laplacian and can be regarded as an extension of it.

%where $\alpha$ is a scaling factor.
% Here we do not follow the conventional definition since the graph is conceptually created to tackle the MIMO detection problem.
% While we specify an $\alpha$ to ensure that $\mathbf{L}_{\text{MIMO}}$ shares spectral properties with the standard graph Laplacian.
% Note that $\lambda(\H^\top \H)>0$. 
% Given that the graph Laplacian is a positive semidefinite matrix with zero as the smallest eigenvalue,
% we take 
% \begin{equation}
% \alpha = 1/\lambda_{\text{max}}(\H^\top \H),  
% \end{equation}
% that lets $\lambda(\I - \alpha \H^\top \H) \in [0, 1)$.
Since in the real-valued system model in equation \eqref{eq:sys-model-real} $\H$ is assumed to have full column rank, multiplying both sides of equation \eqref{eq:sys-model-real} 
by $(\H^\top \H)^{-1} \H^\top$ gives
\begin{align}
\x \!=\! (\H^\top \H)^{-1} (\H^\top\y + \H^\top\n)  
  \!=\! (\alpha \H^\top \H)^{-1} (\alpha \H^\top \y + \alpha\H^\top\n).
\label{eq: sys-model-real-scale} 
\end{align}
Since $|\lambda(\mathbf{L}_{\text{MIMO}})|<1$,
$$
(\alpha \H^\top \H)^{-1}
= (\I-\mathbf{L}_{\text{MIMO}})^{-1}
= \sum_{m=0}^{\infty}  \mathbf{L}_{\text{MIMO}}^m.
$$
Then from \eqref{eq: sys-model-real-scale} we obtain
\begin{align}
\x  
= \Big( \sum_{m=0}^{\infty}  \mathbf{L}_{\text{MIMO}}^m \Big) (\alpha \H^\top \y + \alpha\H^\top\n).
\label{eq: sys-model-graph-convolution}
\end{align}
% We refer to \eqref{eq: sys-model-graph-convolution} as the graph convolution form of the system model.
% Regarding $\alpha \H^\top \y \in \R^{2N_t}$ as the graph signal on ${\cal G}_{\text{MIMO}}$ with some unknown noise,
% \red{Think carefully about this - later on you use the noise information for signal as well.}
This shows that the real-valued system can be modeled as spectral signal convolution on graph, where we have $2N_t$ nodes and weight matrix $\W$. More specifically, $\x$ can be regarded as the graph convolution between the polynomial filter $\g(\lambda(\mathbf{L}_{\text{MIMO}})) = \sum_{m=0}^{\infty} \lambda(\mathbf{L}_{\text{MIMO}})^m$
and the graph signal $\alpha \H^\top \y + \alpha\H^\top\n$ as shown in equation \eqref{eq:gc-poly}.
%and $\x$ is the output of this graph convolution.
Therefore,
a graph convolution-based GNN could be employed to learn the deep representations for each unknown $x_i$ in the system model.

% \red{This is a crucial part. 
% It's not a clear to a reader why you need to do this.
% What's the problem with the existing approach?
% What's the new? What's the significance?}
% \red{More is needed here. What's the innovation? 
% You have to sell your stuff.}
% \red{A reader cannot see this is related to the graph Laplacian, 
% as you don't have a graph yet.}
% \blue{The whole section has been rewrote}

\subsection{Architecture of GCEPNet}
Based on the discussion in previous sections,
we propose a graph convolution-based method, GCEPNet, to improve the estimation of cavity distribution in \eqref{eq:epestimation} with faster inference, which would later be used in the EP iteration.
% For the graph ${\cal G}_{\text{MIMO}}$,
% we use $\mathbf{s}^{(t)}_i$ to denote the hidden representation 
%We use superscript $(t)$
%to denote the hidden representation of each $x_i$ processed by the graph convolution module at 
%for the $t^{th}$ EP iteration.
%which is the input to the proposed convolution-based GNN module.
According to \eqref{eq: sys-model-graph-convolution},
we initialize the input graph signal $\mathbf{S}^{(0)} \in \Rbb^{2 N_t \times 2}$ as
\begin{equation}
\mathbf{S}^{(0)} = [\alpha \H^\top \y, \alpha \sigma_n \H^\top\mathbf{1}],
\label{eq: feature-initial}
\end{equation}
where $\alpha \sigma_n \H^\top\mathbf{1}$ is used to approximate the unknown $\alpha \H^\top \n$ in \eqref{eq: sys-model-graph-convolution},
and the $i$-th row of $\mathbf{S}^{(0)}$ is a graph signal of the node $i$
corresponding to $x_i$.
%for containing the input graph signal and the noise variance are in rows of $\mathbf{S}^{(0)}$.
% \footnote{
% %A remark about \eqref{eq: feature-initial}:
% %Since the noise vector $\n$ in equation \eqref{eq: sys-model-graph-convolution} is unknown, 
% Here $\alpha \sigma_n \H^\top\mathbf{1}$ represents $\alpha \H^\top \n$ in \eqref{eq: sys-model-graph-convolution}. 
% Equation \eqref{eq: input-feature} mitigates this approximation error since $\mathbf{W}_0$ scales $\alpha \H^\top \y$ and
% $\alpha \sigma_n \H^\top\mathbf{1}$ separately.}
% To mitigate this approximation error we separate $\alpha \H^\top \y$ and $\alpha \sigma_n \H^\top\mathbf{1}$ to define $\mathbf{S}^{(0)}$ in equation \eqref{eq: feature-initial} so that the two vectors can be scaled separately later (see equation \eqref{eq: input-feature}).}
% \red{Why not use a random vector following the same distribution as $\n$
% in implementation?}
% \blue{That won't be helpful, we only need the variance information. Personally I think we don't need the second column at all, since GNN is supposed to do de-noising.}
% \red{We can discuss this after the paper is done.}
The key operation in GCEPNet is that we use the graph convolution in equation \eqref{eq: sys-model-graph-convolution} to recover useful information about the unknown $\x$ from the graph signal defined in \eqref{eq: feature-initial}, by  the following approximated Chebyshev polynomials with order $M$,
\begin{equation}
\sum_{m=0}^{\infty} (\mathbf{L}_{\text{MIMO}})^m \approx \sum_{m=0}^{M} w_m T_{m}(\mathbf{L}_{\text{MIMO}}).
\label{eq: sys-model-cheby}
\end{equation}
At $t^{th}$ 
%\red{If you prefer to use $t^{th}$, then change all these.} 
EP iteration,
based on the cavity density of each $x_i$ in \eqref{eq:qti},
we construct $\mathbf{a}^{(t)} = [a_0^{(t)}, \cdots, a_{2 N_t}^{(t)}]^\top$,
$\mathbf{b}^{(t)} = [(b_0^{(t)})^2, \cdots, (b_{2 N_t}^{(t)})^2]^\top$,
and denote $\mathbf{P}^{(t)} = [\mathbf{a}^{(t)}, \mathbf{b}^{(t)}]$, where $\mathbf{P}^{(t)}$ contains the mean and variance information for $x_i$. %under the independent Gaussian assumption in \eqref{eq:iga-2}.
GCEPNet intends to improve the standard cavity density by incorporating pair-wise information learned via the graph convolution on ${\cal G}_{\text{MIMO}}$ efficiently.
In Figure~\ref{fig:GCEPNet}, 
we present the overall structure of GCEPNet.
For $t=0 \dots T$,
%At the $t^{th}$ EP iteration
%where the graph convolution and the GRU module at the $t^{th}$ EP iteration are
\begin{align}
&\bar{\mathbf{S}}^{(0)} = \mathbf{S}^{(0)}  \mathbf{W}_0 + \mathbf{B}_0,
%\quad \text{for} \quad t = 0, 
\label{eq: input-feature}\\
% &\text{For} \quad t = 0, \cdots, T \nonumber \\
&\widetilde{\mathbf{S}}^{(t)} = \text{MLP}_{1} ( \bar{\mathbf{S}}^{(t)} ), \label{eq: hidden-feature}\\
&\hat{\mathbf{S}}^{(t)} = \textstyle \sum_{m=0}^{M} \text{ATT} (\mathbf{S}^{(0)})_m T_{m}(\mathbf{L}_{\text{MIMO}}) \widetilde{\mathbf{S}}^{(t)}, \label{eq: gcepnet}\\
&\bar{\mathbf{S}}^{(t+1)} = \text{GRU}(\hat{\mathbf{S}}^{(t)}, \mathbf{P}^{(t)}),
\label{eq: gru}\\
&\mathbf{P}_{\mathcal{G}}^{(t)} = R (\text{MLP}_{2} (\bar{\mathbf{S}}^{(t+1)}) ).
\label{eq: read-out}
\end{align}
% \sitao{Is $\mathbf{P}_{\mathcal{G}}^{(t)}$ the same as $\mathbf{P}^{(t)}$?}\blue{No}
%\sitao{You should put $t=1 \dots T$ somewhere above}
% \red{You need to explain where this comes from. 
% A reference might be needed. What is the subscript $h$?
% You have to define MLP, ATT, GRU, softmax.
% Will a diagram for this architecture be helpful?}
% \blue{They are expalined in the following three subsections}
% The details of these equations are discussed as follows.
% \subsection{Disentangled Learning}
% %Instead of following the ChebNet as described in \eqref{eq:chebnet} that implicitly learns the spectral coefficients,
% We use separate sets of trainable parameters: one for computing the coefficients of Chebyshev polynomials,
% and others for learning the node hidden representations.
% Where $\alpha \sigma_n \H^\top\mathbf{1}$ is used to approximate the unknown $\alpha \H^\top \n$ in \eqref{eq: sys-model-graph-convolution}. 
% Equation \eqref{eq: input-feature} mitigates the approximation error in \eqref{eq: feature-initial} since $\mathbf{W}_0$ scales $\alpha \H^\top \y$ and
% $\alpha \sigma_n \H^\top\mathbf{1}$ separately.
Equation \eqref{eq: input-feature} is to mitigate the error caused by replacing $\alpha \H^\top \n$ with $\alpha \sigma_n \H^\top\mathbf{1}$ in \eqref{eq: feature-initial}.
At the initial EP iteration,
in \eqref{eq: input-feature},
% The $\text{MLP}_{f}$ is a Multi-Layer Perceptron (MLP) with input size 2 and output size $N_u$,
trainable parameters $\mathbf{W}_{0} \in \Rbb^{2 \times N_u}$ and $\mathbf{B}_0 \in \Rbb^{2 N_t \times N_u}$ transform the initial 2-dimensional graph signal $\mathbf{S}^{(0)}$ to a hidden representation $\bar{\mathbf{S}}^{(0)}$ with size $N_u$.
We use $\text{MLP}$ for a Multi-Layer Perceptron (MLP).
The graph convolution in \eqref{eq: gcepnet}
% \red{This is regarded as graph convolution? 
% Other functions have been mentioned in this paragraph, except MLP.
% Why not define it here so that it won't need to be explained later?}
is applied to the input graph signal $\widetilde{\mathbf{S}}^{(t)}$ to compute the output signal $\hat{\mathbf{S}}^{(t)}$.
The attention-based network $\text{ATT} (\mathbf{S}^{(0)})$ learns an $M+1$ dimensional vector containing data-dependent Chebyshev coefficients $\text{ATT} (\mathbf{S}^{(0)})_m$ for $m=0,1,\ldots,M$.
The gated recurrent unit (GRU) \cite{cho2014learning} in \eqref{eq: gru}
provides a gating mechanism
to integrate the output signal $\hat{\mathbf{S}}^{(t)}$ 
%from the graph convolution in \eqref{eq: gcepnet} 
and %information for 
%the first two moments of 
the %standard 
cavity distribution stored in $\mathbf{P}^{(t)}$. 
The output of the GRU will be the input of the GNN module at the next EP iteration.
The readout module $R$ in \eqref{eq: read-out} computes the mean and variance estimation of each $x_i$.
Two GNN layers perform \eqref{eq: hidden-feature}-\eqref{eq: gru} two times.
More explanations are given in the following two subsections.

\subsection{Data-dependent Graph Convolution Coefficients}
In equation \eqref{eq:chebyshev_polynomial}, ChebNet uses conventional Chebyshev polynomial as graph filter, which uses fixed coefficients $w_m$. However, this architecture is less powerful and cannot perform well on testing data. Thus, in \eqref{eq: gcepnet}, we proposed to replace the fixed $w_m$ by data-dependent learnable attention scores $\text{ATT} (\mathbf{S}^{(0)})_m$.
Training $\text{ATT} (\mathbf{S}^{(0)})$, which is a function
of an instance $\{ \y, \H, \sigma_n \}$, 
can improve the generalization ability of the graph convolution module.
% \sitao{Why does this can improve generalization? Any reference or explanation for the advantage of data-dependent coefficients? Is this a new trick? If yes, it should be one of your contributions.}
% we training an attention-based network $\text{ATT} (\bar{\mathbf{S}}^{(0)})$ instead of directly learning a fixed coefficient vector. 
% Since for MIMO detection,
% we have an inductive graph-level node classification task \cite{hamilton2017inductive},
% where each sample is represented as a graph.
% Although a fixed Chebyshev coefficient vector may fit the training set,
% these coefficients might not perform well on the unseen graphs from the testing set.
% The basic idea to overcome this challenge is that we learn data-dependent coefficients for the graph convolution.
% To be specific,
% we train an attention-based network $\text{ATT} (\bar{\mathbf{S}}^{(0)})$ with the $\bar{\mathbf{S}}^{(0)}$ containing instance information $\{ \y, \H, \sigma \}$  as input and use a calibration vector $\1$ of size $M+1$,
% to calculate the coefficients of the Chebyshev polynomials in \eqref{eq: sys-model-cheby}.
$\text{ATT} (\mathbf{S}^{(0)})$ first computes attention score $\alpha_i$ 
between the default coefficient vector $\1$ with size $M+1$ and each row of the key matrix $\K \in \Rbb^{2 N_t \times (M+1)}$,
%as follows,
\begin{align}
\alpha_i = \frac{\exp (\mathbf{K}_{i, :} \1 )}{\sum_j \exp (\mathbf{K}_{j, :} \1 )},\;\text{where } \mathbf{K} = \text{MLP}_{3}(\mathbf{S}^{(0)}).
\label{eq: atten-feature}
\end{align}
%\red{$\alpha_i$ is a matrix?}
% \sitao{You don't need MLP 1,2,3. Just using MLP is OK.}\blue{I need to refer to them to talk about their size in the complexity analysis.}
The $m$-th column of %the key matrix 
$\mathbf{K}$ is an embedding for the $m$-th coefficient.
The output of $\text{ATT} (\mathbf{S}^{(0)})$ is a weighted average of each dimension of the coefficient embedding in rows of $\mathbf{K}$,
where the attention score $\alpha_i$ serves as the normalized weight.
\begin{equation}
\text{ATT} (\mathbf{S}^{(0)}) = \left(\textstyle\sum_{i} \alpha_i \mathbf{K}_{i, :}\right)^\top.
\label{eq: atten-output}\textstyle
\end{equation}
% \textbf{To our best knowledge, we are the first to introduce data-dependent attention scores in polynomial filters to increase the expressive power of graph convolution networks.}
%\red{It's a row vector}
% \sitao{Why do you learn the weights in this way? Any advantage, explanation or reference? And I don't think it's an attention vector, it's just a weight vector.}
% \blue{We calculate attention score between each row of K and a a calibration vector $\1$ of size $M+1$, where $\1$ can be regarded as the defualt coefficient. We don't have space to elaborate}
\subsection{Readout Module for Graph-based Estimation}
%Suppose the maximum number of EP iterations is $T$,
At $t^{th}$ EP iteration ($t < T$),
the graph convolution-based module gives an improved estimation of the standard cavity distribution in \eqref{eq: read-out}.
The $\text{MLP}_{2}$ maps $\bar{\mathbf{S}}^{(t+1)} \in \Rbb^{2 N_t \times N_u}$ to a $2N_t$ by $2^k$ matrix,
where the $i$-th row gives unnormalized probabilities over the constellation set.
The readout function $R$ applies row-wise softmax normalization in \eqref{eq: softmax}.
Then the $i$-th row of the output $\mathbf{P}_{\mathcal{G}}^{(t)}$
represents the graph convolution-enhanced cavity distribution $q^{(t) \backslash i}_{\mathcal{G}}(x_i) = N(\hat{a}_{i, \mathcal{G}}^{(t)}, (\hat{b}_{i, \mathcal{G}}^{(t)})^2)$ via \eqref{eq: softestimate}, \eqref{eq: varianceestimate}.
% \begin{align}
% \hat{\mu}_{i, \mathcal{G}}^{(t)} &= \sum_{e} \calX_k(e) \times \mathbf{P}^{(t)}_{\mathcal{G}}(i, e) \label{eq: gnn-mean}\\
% (\hat{\sigma}^{(t)})^2_{i, \mathcal{G}} &= \sum_{e} (\calX_k(e) - \hat{\mu}_{i, \mathcal{G}}^{(t)})^2 \times \mathbf{P}^{(t)}_{\mathcal{G}}(i, e),
% \label{eq: gnn-variance}
% \end{align}
which is used to replace the cavity distribution $q^{(t) \backslash i}(x_i)$ 
in \eqref{eq:epestimation} to compute the posterior $\hat{p}^{(t)}(x_i)$, 
and later to update the parameters for EP in \eqref{eq:ep-update}.
At the last EP iteration ($t = T$),
GCEPNet output the detection $\hat{x}_{\mathcal{G}}$ %$\x_{\mathcal{G}}$ 
with $\mathbf{P}_{\mathcal{G}}^{(T)}$ in \eqref{eq: round}.
% computes the final estimation $\hat{\x}_{\mathcal{G}}$ for each $x_i$ via \eqref{eq: gnn-mean} with $\mathbf{P}_{\mathcal{G}}^{(T)}$,
% and get the hard detection $\x_{\mathcal{G}}$ by \eqref{eq: round}.

% \subsection{Training}

% \begin{equation}
% \mathcal{L} = 
% \end{equation}

% \red{Is it common to define the loss function in this experimental section not this section?}
% \blue{It is okay I think.}

\section{Computational Complexity Analysis}
% \red{This section is not complete?}
% Comparing with GEPNet,
% which also uses a GNN module to improve the cavity distribution,
\label{sec:gepnet-compare}

% The system-model inspired graph convolution in the proposed GCEPNet significantly improves the efficiency of the spatial-based GNN in GEPNet,
% thus scales better with the problem size. 
% \red{You want to give a conclusion before analysis?}
%by the system model inspired graph convolution.
% The following discussion assumes that GCEPNet and GEPNet use the same hidden size.
%we show that GCEPNet scales better with the problem size than GEPNet.

% \begin{figure}
% \centerline{\includegraphics[width=0.5\textwidth]{fig/compare_inference_time_b10_iter5000.png}}
% \caption{The total inference running time (in seconds) for $5 \times 10^4$ samples with $N_t = N_r$ and the same number of EP iteration ($T=10$). All methods are under CUDA acceleration}  
% \label{fig:time_comparison}
% \end{figure}

%\subsection{Efficiency Comparisons with GEPNet}

% \begin{equation}
% \s_i^{(0)} = [\y^\top \h_i, \h_i^\top \h_i, \sigma^2], \quad \boldsymbol{\epsilon}_{ij} = [-\h_i^\top \h_j, \sigma^2]
% \end{equation}
This section compares the complexity of the GNN module in GCEPNet and GEPNet.
Both GCEPNet and GEPNet introduce additional computations to EP,
which originate from their GNN modules.
In \eqref{eq: hidden-feature} and \eqref{eq: gcepnet},
GCEPNet uses the graph convolution to transform the input graph signal $\bar{\mathbf{S}}^{(t)}$ to the output signal $\hat{\mathbf{S}}^{(t)}$.
GEPNet uses a less efficient GNN to achieve this purpose.
% Denote $\bar{\s}_i^{(t-1)} \in \Rbb^{N_u}$ as the hidden representation for $x_i$ at the $t^{th}$ EP iteration,
The spatial GNN in GEPNet specifies the feature of the edges $e_{ij}$ as $\boldsymbol{\epsilon}_{ij} = [-\h_i^\top \h_j, \sigma^2]$,
and is characterized by the following aggregation
\begin{equation}
\hat{\s}_i^{(t)} = \sum_{j \neq i} \text{MLP}_{4} \left([\bar{\s}_i^{(t)}, \bar{\s}_j^{(t)}, \boldsymbol{\epsilon}_{ij}] \right).
\label{eq: gepnet-gnn}
\end{equation}
We assume that those $\text{MLP}$ used in GEPNet and GCEPNet all have two hidden layers with sizes $N_{h_1}$ and $N_{h_2}$,
to be consistent with \cite{kosasih2022graph}.
The $\text{MLP}_{4}$ in \eqref{eq: gepnet-gnn} has an input layer with size $2N_u + 2$ and an output layer with size $N_u$.
Denote $N = 2N_t$ for simplicity,
then the total cost for computing the updated hidden representations for all $x_i$ 
via \eqref{eq: gepnet-gnn} is $\mathcal{C}_{1} =  N(N-1)[(2N_u + 2)N_{h_1} + N_{h_1} N_{h_2} + N_{h_2} N_u + N_u]$.
Under the setting in \cite{kosasih2022graph} that $N_u = 8$, $N_{h_1} = 64$, $N_{h_2} = 32$, 
the coefficient of $N^2$ is 3464 and is not negligible.
% It indicates that
% the computation complexity of the GNN in GEPNet is in $\mathcal{O}(N^4)$ when,

\begin{figure}[t]
\centerline{\includegraphics[width=0.5\textwidth]{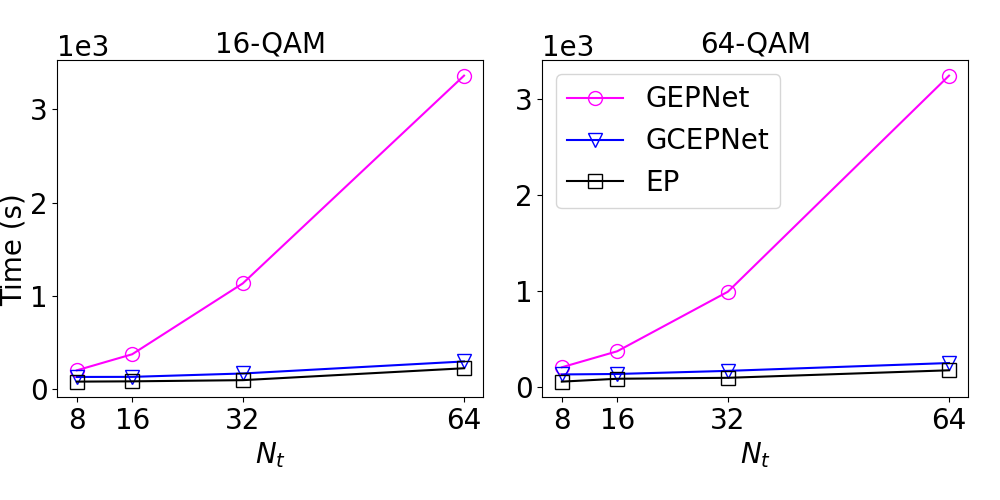}}
\caption{The total inference running time (in seconds) for $5 \times 10^4$ samples with $N_t = N_r$ and the same number of EP iterations ($T=9$). 
Both GCEPNet and GEPNet use 2 GNN layers in each EP iteration.
The implementation of GEPNet is from \cite{kosasih2022graph}.
All methods are under CUDA acceleration.}  
\label{fig:time_comparison}
\end{figure}

For GCEPNet,
the $\text{MLP}_{1}$ in \eqref{eq: hidden-feature} has an input and an output layer of size $N_u$,
and the $\text{MLP}_{3}$ in \eqref{eq: atten-feature} has an input of size $2$ and an output of size $M+1$,
% two hidden layers with size $N_{h_1}$ and $N_{h_2}$, and a output layer with size $N_u$
then the total cost for the graph convolution in \eqref{eq:L-MIMO}, \eqref{eq: hidden-feature}, \eqref{eq: gcepnet}, \eqref{eq: atten-feature} and \eqref{eq: atten-output} is $\mathcal{C}_{2} = N(N_u N_{h_1} + N_{h_1} N_{h_2} + N_{h_2} N_{u} + N_{h_2} M + M) + N^2N_uM$.
When $M<433$,
$\mathcal{C}_{2}$ has a smaller coefficient of $N^2$ than $\mathcal{C}_{1}$.
%$\mathcal{C}_{2} < \mathcal{C}_{1}$.
% With a moderate $M$,
% $\mathcal{C}_{2}$ is much smaller than $\mathcal{C}_{1}$.
% still $\mathcal{O}(N^2)$ 
% \red{Do you mean a moderate $M$? If $M$ is large, $C_1$ is large than $C_2$?}
In practice, we use $M=3$.
Figure~\ref{fig:time_comparison} demonstrates that,
comparing with EP,
GEPNet\,\footnote{Following the implementations in \cite{kosasih2022graph}: https://github.com/GNN-based-MIMO-Detection/GNN-based-MIMO-Detection} does not scale with the problem size as a result of the inefficient GNN aggregation,
while GCEPNet effectively resolves the bottleneck with the newly proposed graph convolution.

%we plot the total inference time for $5\times10^4$ samples of GEPNet and GCEPNet.
% \red{Does Fig 2 just give the comparison of the cost of one iteration
% for one set of chosen parameters (i.e., the dimensions?)?
% Why not given the cost comparisons for the same data given in 
% the next section?}

\section{Numerical Results} \label{sec:exp}
% \red{Did you explain why do comparison with GEPNet and EP only?}
% \blue{Is that already clear based on what we said in the introduction?}
This section mainly compares the detection performance of GCEPNet with EP and GEPNet.
GEPNet outperforms other detectors \cite{kosasih2022graph2} except IEP-GNN \cite{liu2023model},
which has the same complexity as GEPNet.
Our GNN module can be effortlessly combined with the EP module of 
IEP-GNN and hopefully the combined one can achieve the best performance and efficiency. 
But because IEP-GNN is not open-sourced, our numerical tests will not involve it.
In experiments,
entries of $\H_c$ are generated from \iid{} $\mathcal{CN}(0, \omega_c^2)$ with $\omega_c^2 = 1/N_r$ then get $\H$.
The $\x$ is uniformly sampled from $\calX_k^{2N_t}$ with $\sigma_\x^2=(4^k-1)/3$.
We find $\sigma_n$ 
%for generating $\mathbf{n}$ 
for the signal-to-noise ratio ($\text{SNR}_{\text{dB}}$) defined by:
\begin{equation}
\text{SNR}_{\text{dB}} = 10 \log_{10} (\Exp{\| \H_c \x_c \|_2^2}/\Exp{\| \n_c \|_2^2})
%= 10 \log_{10} \frac{2N_t(4^k - 1)\omega_c^2}{3\sigma_c^2},
\label{eq: snr}
\end{equation}
%\red{So $\x$ is uniformly distributed?}
%\blue{Yes}
Then we generate $\mathbf{n}$ and calculate $\y$.
We use $\{\y, \H, \sigma_n\}$ as the network input, and $\x$ for computing the network loss and the symbol error rate (SER). 
Let $\Z \in \Rbb^{2N_t \times 2^k}$ be the label matrix with $\Z(i, c) = 1$ if $x_i = \calX_k(c)$ and $0$ otherwise,
the sample loss is defined as:
\begin{equation}
\mathcal{L} = - \text{trace}(\Z^\top \ln \mathbf{P}_{\mathcal{G}}^{(T)}).
\end{equation}
%\red{Can you look at the GEPNet paper to see if the definition of SNR is the same one}
%\blue{The definition of SNR is the same, but they have a different setting where they normalize the constellation set so the variance of $x$ changes.}
% Both training and validation samples are generated with $\text{SNR}_{\text{dB}} \in [20, 60]$.
%With the average sample loss,
The training uses 850 epochs to optimize parameters in each network to minimize the average sample loss.
There are 100 iterations per epoch and each iteration contains 100 training samples.
Every training sample has a $\text{SNR}_{\text{dB}}$ uniformly sampled from $[25, 50]$.
In each epoch,
the network is tested on a validation set containing 2000 samples for each $\text{SNR}_{\text{dB}}$ from $\{25, 26, \cdots, 50 \}$.
%and use the mean SER over all samples as the performance criteria.
The training uses the Adam optimizer with a learning rate of $10^{-3}$,
which is adjusted very 100 epochs according to the validation loss by the ReduceLROnPlateau scheduler in PyTorch.
The network with the best mean SER on the validation set is selected for the testing stage.
All networks are trained for each problem size and QAM configuration.

\begin{figure}[t]
\centerline{\includegraphics[width=0.5\textwidth]{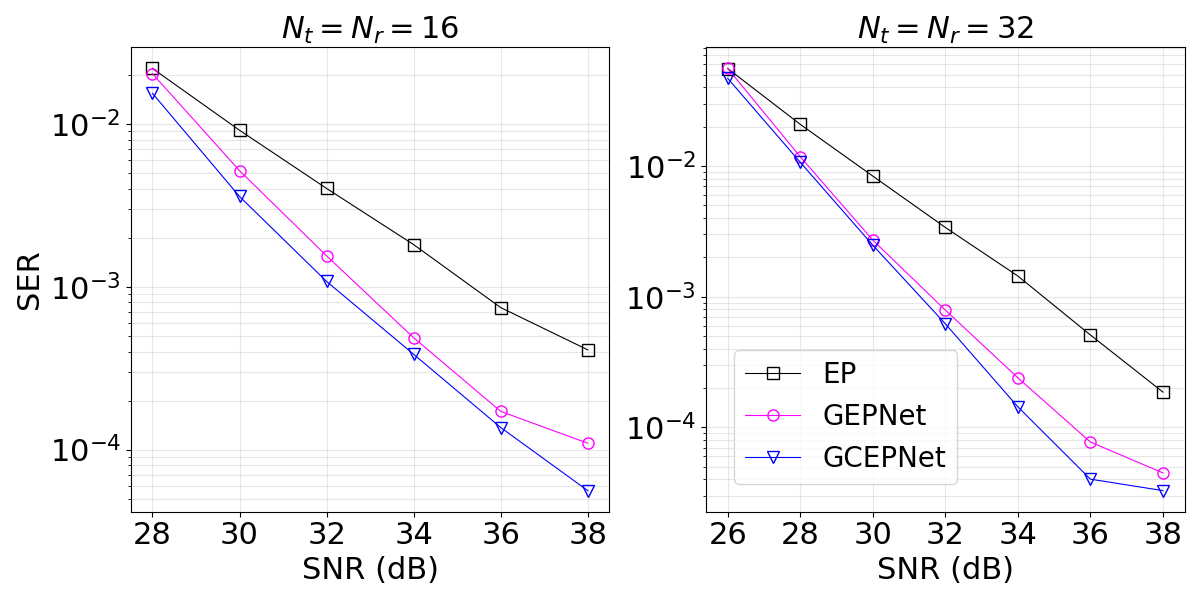}}
\caption{The SER performance comparison for 64-QAM}  
\label{fig:results}
\end{figure}

The testing uses $10^5$ samples for each $\text{SNR}_{\text{dB}}$.
Figure~\ref{fig:results} compares the average SER of GCEPNet, GEPNet \cite{kosasih2022graph} and EP \cite{cespedes2014expectation} for $N_t = N_r = 16$ and $N_t = N_r = 32$ with 64-QAM.
Both GCEPNet and GEPNet enhance the performance of EP by incorporating correlation information into the standard cavity estimation. 
While GCEPNet is significantly more efficient than GEPNet,
it consistently outperforms GEPNet as well.
Given that GEPNet achieves the best performance among existing sub-optimal detectors \cite{kosasih2022graph2},
GCEPNet is established as the new state-of-the-art (SOTA) method.

% \begin{figure}
% \centerline{\includegraphics[width=0.5\textwidth]{fig/16x16_k3_random.png}}
% \caption{$N_t = N_r = 16$, 64-QAM}  
% \label{fig:results_16}
% \end{figure}

% \begin{figure}
% \centerline{\includegraphics[width=0.5\textwidth]{fig/32x32_k3_random.png}}
% \caption{$N_t = N_r = 32$, 64-QAM}  
% \label{fig:results_32}
% \end{figure}

\section{Conclusion}

\label{sec:conclusion}

We have proposed GCEPNet,
a graph convolution-enhanced expectation propagation detector.
Our analysis and numerical tests have demonstrated that GCEPNet
significantly improves the efficiency of the state-of-the--art GNN-based counterparts,
and consistently outperforms the existing detectors.
% \bibitem{b1} 

% \end{thebibliography}
\bibliography{ref.bib} 
\bibliographystyle{./IEEEtran}

\end{document}